\documentclass[sn-mathphys]{sn-jnl}% Math and Physical Sciences Reference Style
\usepackage[justification=centering]{caption}
\usepackage{lineno,hyperref}
\modulolinenumbers[5]
\usepackage{verbatim}
\usepackage{float}
\usepackage{xcolor}
\usepackage{amsmath}
\floatstyle{plaintop}
\restylefloat{table}
\jyear{2021}%

\theoremstyle{thmstyleone}%
\theoremstyle{thmstyletwo}%

\theoremstyle{thmstylethree}%

\begin{document}

\title[Weakly-Supervised Action Localization, and Action Recognition]{Weakly-Supervised Action Localization, and Action Recognition using Global-Local Attention of 3D CNN}


\author*[1]{\fnm{Novanto} \sur{Yudistira}}\email{yudistira@ub.ac.id}

\author[2]{\fnm{Muthu Subash } \sur{Kavitha}}\email{kavitha@nagasaki-u.ac.jp}
\equalcont{These authors contributed equally to this work.}

\author[3]{\fnm{Takio} \sur{Kurita}}\email{tkurita@hiroshima-u.ac.jp}
\equalcont{These authors contributed equally to this work.}

\affil*[1]{\orgdiv{Informatics Engineering}, \orgname{Faculty of Computer Science, Brawijaya University}, \orgaddress{\street{Veteran st. 8}, \city{Malang}, \postcode{65145}, \state{East Java}, \country{Indonesia}}}

\affil[2]{\orgdiv{School of Information and Data Sciences}, \orgname{Nagasaki University}, \orgaddress{\street{1-14 Bunkyo-machi,}, \city{ Nagasaki}, \country{Japan}}}

\affil[3]{\orgdiv{Graduate School of Advanced Science and Engineering}, \orgname{Hiroshima University}, \orgaddress{ \city{Higashi-hiroshima}, \postcode{739-8521}, \state{Hiroshima}, \country{Japan}}}


\abstract{3D Convolutional Neural Network (3D CNN) captures spatial and temporal information on 3D data such as video sequences. However, the information loss that occurs seems unavoidable due to the convolution and pooling mechanism. To improve the visual explanations and classification in 3D CNN, we propose two approaches; i) aggregate layer-wise global to local (global-local) discrete gradient using trained 3DResNext network, and ii) implement attention gating network to improve the accuracy of the action recognition.
The proposed approach intends to show the usefulness of every layer termed global-local attention in 3D CNN via visual attribution, weakly-supervised action localization, and action recognition. Firstly, the 3DResNext is trained and applied for action classification using backpropagation concerning the maximum predicted class. The gradient and activation of every layer are then up-sampled. 
Later, aggregation is used to produce
more nuanced attention points out the most critical part of the predicted class’s input videos. We use contour thresholding of final attention for final localization. We evaluate spatial and temporal action localization in trimmed videos using fine-grained visual explanation via 3DCAM. Experimental results show that the proposed approach produces informative visual explanations and discriminative attention. Furthermore, each layer's action recognition via attention gating produces better classification results than the baseline model.}

\keywords{3D CNN, spatio-temporal attribution, visual attribution, weakly-supervised localization, action recognition}



\maketitle

\section{Introduction}\label{sec1}

Methods based on deep neural networks (DNNs) have achieved impressive results for several computer
vision tasks, such as image classification, object detection, and image generation. Combined with
the general tendency, the DNNs focus on
high quantitative performance. Hence the computer vision community has been motivated to use wide adoption of DNN-based methods, despite
the initial skepticism due to their black-box characteristics. To improve the explanations, we aim to visualize the model’s visual descriptive predictions by bridging the gap between the learned features and its explanations regarding its decision. 
Several CNN-based tasks have been developed with visual descriptions, producing high qualitative and quantitative performance. The recent trend of action recognition or visual explanation tasks uses end-to-end learning via deep CNN based on a large-scale dataset. Moreover, CNN’s architecture is the chain of convolution with pooling to capture the fine-grained information local to global fashion. Despite their superiority, there is a question about making use of each layer to understand
classification results better and improve recognition based on it. Significantly, the 3D CNN can effectively preserve the spatiotemporal information inside the structure. Moreover, in the ResNet family \cite{he}, the last layer tends to visualize the non-detailed information due to residual connections making its weights denser regardless of its capability to avoid vanishing and exploding gradients.

In the popular GradCAM \cite{selv} attribution method, the last layer is used to target the visual explanation. Consequently, the attention is not specific and less sensitive than the non-residual network like VGGNet \cite{simonyan}. The global-local depth with non-linearity inside the ResNet family is also considered meaningful because of its better local minima than the shallower network’s global minima \cite{kawaguchi}. However, it is still unclear how to visualize explainable classification attention in trained residual 3D CNN and its importance for the classification task. To investigate this, we use two approaches. One approach uses the trained 3DResNext \cite{xie} network and visualizes the attribution of all the layers except classification layer. Another approach is implementing a layer-wise attention gating network to recognize action with Inception 3D (I3D). 

The visual explanation by attribution is critical method to investigate whether the trained network has suitably learned labeled data or not and understand where the network attention is located. For example, it is useful to understand the most supportive region for the decision in medical diagnosis. In spatiotemporal space with 3D CNN, understanding the location of the attention area is of high interest. Motivating this, in this study, we consider generating a network with an accurate visual explanation to interpret the action location for action classification.

Attribution is often called weak localization by performing the attention localization via backpropagation using convolutional layers based on specific labels. Several works have been developed to visually explain the networks. Oquab et al. used multiscale networks rather than up-sampling series from one network for object localization \cite{oquab}. Bazzani et al. \cite{bazzani} developed self-taught object localization by exploiting clustering with a trained classification network. Zeiler and Fergus \cite{zeiler} introduced a visualization method using a     deconvolution layer to map the features back to the pixels. 

The more recent and more straightforward approach is GradCAM \cite{selv}. It visualizes salient attribution of detected objects ranging from classification, image captioning to visual question answering using gradients and activation. The visualization generated by GradCAM tends to point out general attention due to the spatio-temporal pooling of the residual 3D CNN.  Correspondingly, in action understanding, it should be able to visualize such salient spatio-temporal action. However, this approach is still limited. It can not capture temporal attribution for 3D sequence input.

In 2D architecture, similar to GradCAM,  the class activation mapping (CAM) \cite{beng} approach is used for localization. It changed the last fully connected layers with convolution layer and followed with global average pooling, max pooling and log-sum-exp pooling. However, it cannot cover the temporal information, which is similar to the GradCAM approach. Moreover, it only considers unique architecture with global average pooling before softmax. 
Another approach using aggregation of gradients of interpolation between baseline and input points delivered a more proper attribution was observed by visually better sensitivity than the existing methods \cite{sundararajan}. However, these 2D architectures can not capture temporal information and overall fine-grained visualization.

To overcome this, we introduce a new approach that combines the feature maps using the gradient signal that does not require any modification in the network architecture. Furthermore, merging all layers in a global-local fashion helps to enhance the sensitivity and interpreted-ability. In action localization, it is necessary to answer the question, whether the 3D CNN structure truly defines the action over the temporal space or not. This could help researchers to understand better the explanation inside the trained network of the residual 3D CNN. Generally, high sanity for visual explanation is provided that measures the sensitivity and effectiveness of any instances \cite{adebayo}.

Another important challenge is how to evaluate the attribution. In order to evaluate this, several approaches used sanity check using cascading randomization or data randomization \cite{adebayo}. They consistently showed that the GradCAM sensitively changes the attention when the model capability decreases or the model uses randomized labels to train. Such sensitivity measure confirms the sanity check of the attribution and thus succeeded in passing the testing. Attribution is also said to be a weakly supervised task due to the information about the location is not provided, though it is previously trained using annotated datasets. Our approach answers all the above issues and can be applied to any CNN-based or 3D CNN-based architectures. The first approach is to use a trained 3DResNext network and visualizes all the layers’ attribution before pooling. It uses layer-wise normalization and thus provides more precise and more refined visual explanations.
Furthermore, it can achieve localization in one shot and fine-grained fashion. It means it only requires a single forward, a partial backward pass, and layer-wise sequential up-sampling per sequence of frames followed by aggregation. Thus it can effectively capture spatiotemporal attribution. 

Another approach is implementing an attention gating network by recognizing the action into I3D network in each layer before pooling. We also use the gating version of 3D ResNext architecture for action classification. Each layer output after the pooling consists of activation visualization that can be represented as attention. Such an approach is well known as attention gated in medical applications \cite{schlemper}. However, the more general capability of such a 3D action recognition strategy is essential to confirm the attention. Hence this study aims to prove experimentally and show every residual layer in a global-local manner that generalization performance increases. Furthermore, it increases the generalization performance in the visual attribution and strengthens the global-local residual attention for action classification.

Thus, the contribution of this study can be summarized as follows:\\
(1) We show the importance of global-local attention of trained residual 3D CNN architecture in visual attribution, weakly-supervised action localization and action recognition. \\
(2) Propose 3DCAM, the high sensitivity visual attribution to capture spatio-temporal attention for a visual explanation of 3D CNN.\\
(3) Implements attention gating network to recognize actions by exploting layer-wise gates for action classification.\\
(4) Evaluates the visual attribution by comparing the proposed approach with the baseline visual attribution, GradCAM, on different datasets.\\
(5) We compare the action recognition of our gated approach with the baseline architecture for action classification. 

\section{Related Works}

Previously, 3D CNN was found to be challenging to achieve state-of-the-art in action classification due to the lack of video data compared to 2D CNN (since the availability of ImageNet dataset \cite{deng}). With the help of big datasets such as the Kinetics dataset \cite{carreira} with extensive video data, 3D CNN is proved to be comparable to 2D CNN with spatial and flow information with Imagenet pre-train \cite{yudistira} \cite{hara}. Moreover, by visualizing the attention inside a network, it is necessary to know which part of the neuron contributes to the decision. The attention information generated from the network benefits visual explanation, and generalization results in the related literature \cite{yudistira} \cite{hara}. Therefore, this study conducts experiments based on visual attribution and action classification. We divide the related works into two subsections: visual explanation and action recognition.

\subsection{Visual Explanation}

 Visualization of intermediate activation in the form of attention makes humans easier to interpret the classification results. However, it is necessary to investigate the deep learning black box, which consists of several layers and activation functions.  It is interesting to answer whether deep CNN can also behave like humans in object inference. The novel effort of layer-wise attention in CNN is developed for image classification \cite{zeiler}. Based on activation layers, it is found that the network trained on ImageNet can be transferred to the network that trained on the smaller dataset.

 An important, challenging question is whether the 3D CNN can also be treated in visual attribution that consists of spatio-temporal information, similar to 2D CNN. The use of backpropagation by GuidedCAM, GradCAM and GradCAM ++ \cite{aditya} is shown to be meaningful for visual attribution of deep networks and proved in several networks \cite{adel} \cite{tran}. Adel et al. visualize RNN attribution in spatio-temporal space using excitation backpropagation \cite{adel}. The well known C3D network for capturing spatiotemporal features for action classification reveals that the earlier frame tends to capture spatial and temporal and then gradually focus on the temporal part in the consecutive frames  \cite{tran}. Fukui et al. \cite{fukui} introduced an attention branch that learns salient regions within the input image and showed better performance than CAM and GradCAM. It also improves the generalization of image classification and facial attribute recognition. However, it used a 2D network based architecture starting from the input image to output. Therefore, it is interesting to investigate how such explanations can be adopted in 3D CNN for action recognition.

Integrated gradients (IG) was introduced to generate more nuanced attribution and improve sensitivity by interpolating baseline with input feature \cite{sundararajan}. It guarantees several properties like sensitivity, implementation invariant, and linearity with which attribution is trusted. However, based on the sanity check held by Google Brain \cite{adebayo}, IG failed to produce smooth attribution. Whereas, GradCAM smoothly produced the correlation by using a weaker network with random labels. Thus it can be assumed that new methods derived from GradCAM are more explainable for a more sophisticated and adaptable network. 

\subsection{Action Recognition}
Several methods have been developed based on attention gated for action recognition. Attentional pooling for action recognition improved recognition accuracy \cite{girdhar}. It forced the network to learn the specific parts of the image relevant to the task, similar to the weighted pooling. Furthermore, the attention-based network focuses only on the salient regions of the medical image \cite{schlemper}. In image captioning,  \cite{chen} shows the importance of attention generated from the multi-layer and channels-wise last convolution layers. Attention also benefits person re-identification by introducing global and local branches in which the attention selection mechanism was introduced \cite{li}. It proved the importance of local attention in network architecture. Our concerns are similar to those methods in which we exploit local attention on visualization and action recognition on residual 3D CNN. The attention branch \cite{fukui} was recently incorporated to improve accuracy while constructing better visual explanations by learning the feature extractor’s
output. This study assumes that similar information is already available in global-local residual layers by directly attaching the classification layer to each.       

\section{Method}
\subsection{Problem Setting}

The problem of full attribution can be illustrated as follows:

Let us consider deep learning function $F$, where $x$ is the input with $j$ dimensions, and $F: R^j \rightarrow [0,1]$. The attribution $(\beta_1,...,\beta_j)$ is the contribution of the pixel $(x_1,...,x_j)$ to $F(x)$. Moreover,  $(l_1,...,l_z)$ indicates $z$ convolution layers (feature maps) and $(\alpha_1,...,\alpha_z)$ indicates $z$ activation maps inside the deep network. Hence $\beta_{ik}$ is the contribution of the pixel $x_i$ in the convolution layer $l_k$ to $F(x)$ where $i$ and $k$ are index of pixels and layers, respectively. Note that even though the backpropagation is done discretely for each layer, chained gradients of target label with respect to each layer are involved. Thus, one layer’s gradient depends on another layer; it can help to find the instance categories of images and understand the label’s contributed pixels. This property is essential for investigating the interpretation of the network.

Finally, we implement two approaches. One approach uses the trained 3DResNext \cite{xie} network and visualizes the attribution. It shows the importance of global-local attention in terms of visual attribution. Another approach is implementing an attention gating network by recognizing the action with I3D in each layer before pooling.
We also use 3D ResNext to be gated with existing architecture for action classification.

\begin{figure*}[!htb]
\centering
\includegraphics[width=1.0\textwidth]{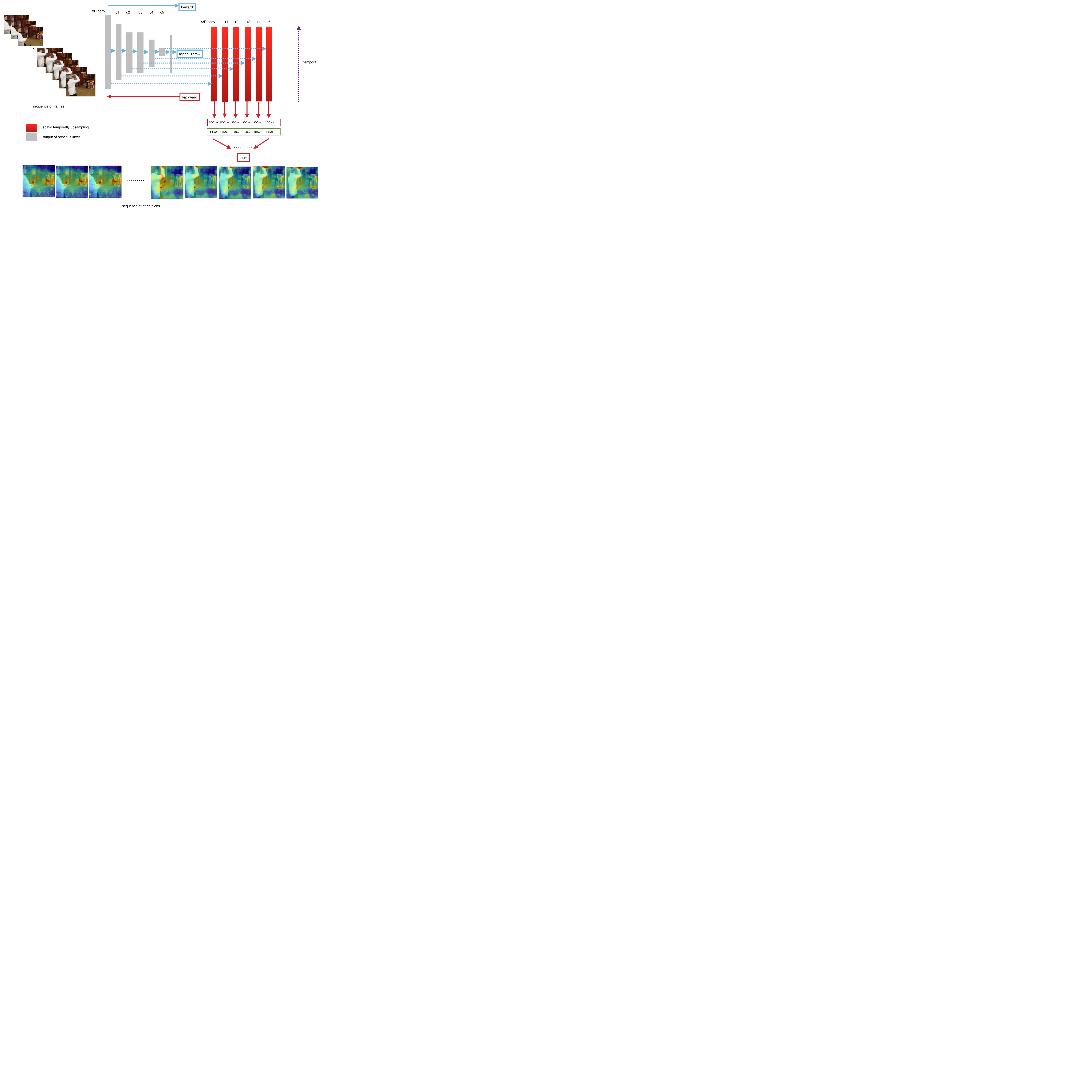}
\caption{\label{fig:3dCAM} Proposed spatiotemporal visual attribution for action recognition on the sequence of frames. Network classifies clips and then back-propagates based on the maximum class score to highlight salient parts. The up-sampled local attentions are pooled to produce the final visualization. The network highlights the hand and objects to be thrown over time in the sequence of throwing actions.}
\end{figure*}

\subsection{Proposed Visual Attribution Network via 3DCAM}
Recently, deep learning works suggested that more profound architecture benefited visualization \cite{zeiler}\cite{oquab}. On the other hand, recent visual attribution-based methods recommended that only the last layer captures coarse localization, especially on the residual network \cite{selv}. Inspired by these, this study aggregates fully global to the local (global-local) discrete gradients of trained networks for action recognition. Consider all layer-wise attributions is $ (\beta_{1},...,\beta_{k})$ and thus global-local attribution can be expressed using layer-wise aggregation of $\sum_{k=1}^{n} \beta_k$.  On the other hand, IG considers the straight line path (in $R^d$) from the baseline $x_0$ to the input $x$, and computes the gradients at all points along the path. We can assume that $x_0 \in R^d$ is zero-valued pixels in black image and $x \in R^d$ is original pixels in input image. Integrated gradients are obtained by accumulating these gradients. Specifically, integrated gradients are defined as the path integral of the gradients (derivative of $F(x_0+\gamma(x-x_0))$ w.r.t. $x_m$ where $m^{th}$ is $th$-dimension and $\gamma$ is weight) of image along the straight line path from the baseline $x_0$ to the input $x$. Unlike IG, which interpolates between input and baseline, this approach interpolates vertically between the top and bottom layers.

 In this work, temporal localization is investigated using gradient information of trained residual 3D CNN (ResNext) based on the maximum decision score for the classification of action recognition (Figure \ref{fig:3dCAM}). Firstly, it explains the architecture that follows forward propagation through the number of layers until the predicted class with maximum value is determined. Secondly, concerning the maximum class score, all layers' layer-wise gradients are obtained by discrete backpropagation, which finds layer-wise salient weights. These weights include attribution information of the predicted class over the spatiotemporal space. Thirdly, each layer's gradient and activation field is upsampled using the same spatiotemporal size with the output sequence (red color). Finally, 3DCAM on each layer is applied before the final aggregation and nonlinear activation function (ReLU), which localizes the actions.

As shown in Figure \ref{fig:3dCAM}, firstly, the network is trained, and the gradient weights $\beta$ are obtained by backpropagating with respect to the maximum class score. The result is then incorporated with ReLU activation to omit negative gradients and avoid false positive attentions. Finally, we set the class that can obtain a maximum score using the following equation:

\begin{equation}\label{sumchannel}
\beta_k=
  \sum_{c}{ \frac{\partial max(y)}{\partial \alpha_{ck}}}
\end{equation}

Where $c$, $k$, $\alpha$, and $y$ indicate the index of channels, selected layers, activation field, and class scores, respectively. Then the parameters $\beta$ and $\alpha$ are upsampled using the trilinear interpolation. It is an extension of linear and bilinear interpolation to treat cubic 3-dimensional enlargement. After upsampling, the parameters $\beta$ and $\alpha$ are normalized by using its maximum value, which is fed into the nonlinear operation before multiplying with each other. It generates 3DCAM of layer $k$, which is defined as:

\begin{equation}\label{3DCAM}
3DCAM_{k} = \textit{ReLU}\bigg(\frac{
    \begin{array}[b]{r}
      \uparrow\beta_k
    \end{array}
  }{
    max( \uparrow\beta_{k})
  }
  )\bigg)
  \textit{ReLU}\bigg(\frac{
    \begin{array}[b]{r}
      \uparrow \alpha_k
    \end{array}
  }{
    max( \uparrow \alpha_k)
  }
  )\bigg)
\end{equation}

Where $\uparrow$ is upsampled tensor, this study's multiplication of gradient and activation is similar to the GradCAM in selecting features based on their saliency.

\subsection{Trilinear Interpolation}

One of the fast and simple methods for upsampling is trilinear interpolation \cite{preim}. Denote $L, B, T$ are linear, bilinear, and trilinear interpolation, respectively, as shown in the equation \ref{linear}, \ref{bilinear}, and \ref{trilinear}, respectively. We start from the most basic linear interpolation, which interpolates the position of current points in 1-dimensional space. Weight $a$ is the current point position normalized into the range of $0 \leq a \leq 1$. target corner point of $v_i$ contain the target size at position index of $i$ (either 0 or 1), such that:

\begin{equation}\label{linear}
L(a) = v_{0}(1-a) + v_{1}a \rvert 0\leq a \leq 1\\
\end{equation}

Then the result of linear interpolation (L) is used to calculate a higher interpolation of bilinear interpolation (B). Target point $(v_{ij})$ is in a 2-dimensional space where i,j are the indexes of the first and second dimensions, respectively. $a_2$ is the weight on the second dimension as formulated as follows:

\begin{equation}\label{bilinear}
\begin{split}
B(a_1, a_2) = (v_{00}(1-a) + v_{10}a))(1-a_2)\\
            +(v_{01}(1-a) + v_{11}a))a_2
\end{split}
\end{equation}

Finally, the trilinear interpolation $T(a_1,a_2,a_3)$ with $a_1$, $a_2$, and $a_3$ as the weight on the first ($p$), the second ($q$), and third ($r$) dimension index, respectively is abbreviated to show interpolated points in 3-dimensional space. $v_{pqr}$ is target corner points on 3-dimensional space ($p$,$q$,$r$). For example, target corner points of $v_{000}=(0,0,0)$ and $v_{111}=(224,224,224)$ are the final size of the enlargement performed at the index position of (0,0,0) and (1,1,1).

\begin{equation}\label{trilinear}
\begin{split}
 T(a_1, a_2,a_3) = (((v_{000}(1-a_1) + v_{100}a_1))(1-a_2)\\
                    +((v_{010}(1-a_1) + v_{110}a_1))a_2)(1-a_3)\\
                    +(((v_{001}(1-a_1) + v_{101}a_1))(1-a_2)\\
                    +((v_{011}(1-a_1) + v_{111}a_1))a_2)a_3\\
\end{split}
\end{equation}

Trilinear interpolation looks for neighboring vertices in 3-dimensional space (volume). However, we often need to calculate sample points within a volume cell. In the early era of volumetric data, this was achieved by interpolation between nearest neighbors. However, like trilinear, the nearest neighbor interpolation produces low visual quality due to intermittent in the middle between adjacent voxels, resulting in the appearance of upsampled blocks. The results can be improved by utilizing Gaussian filtering to refine the interpolation.

\subsection{Aggregation of Global-local Explanations  }

\begin{figure*}
\centering
\includegraphics[width=0.9\textwidth]{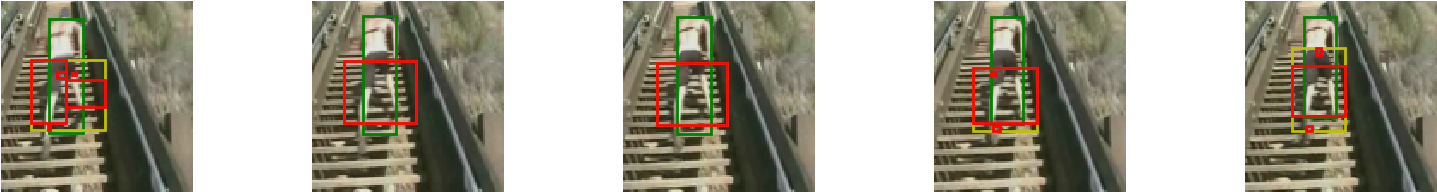}\\
\includegraphics[width=0.9\textwidth]{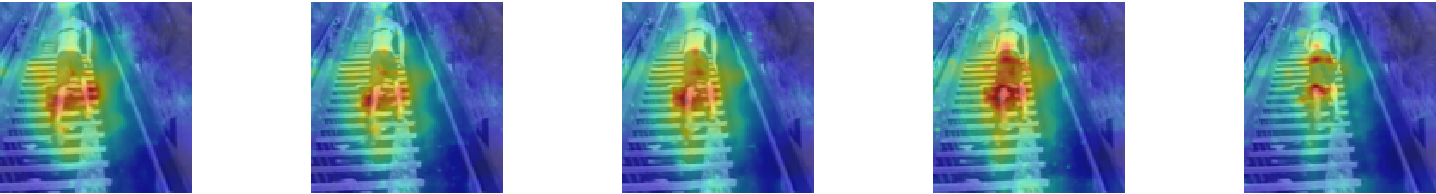}\\~\\
\includegraphics[width=0.9\textwidth]{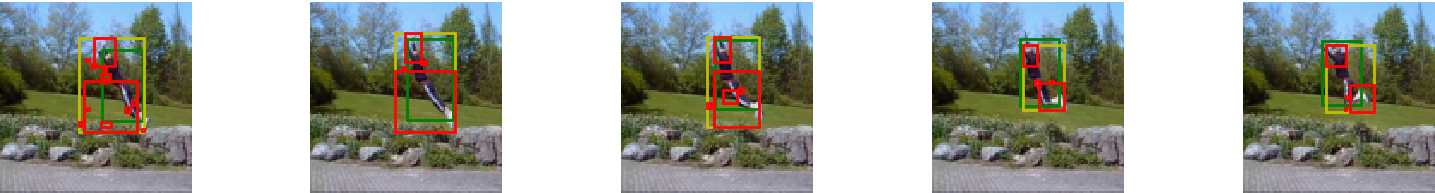}\\
\includegraphics[width=0.9\textwidth]{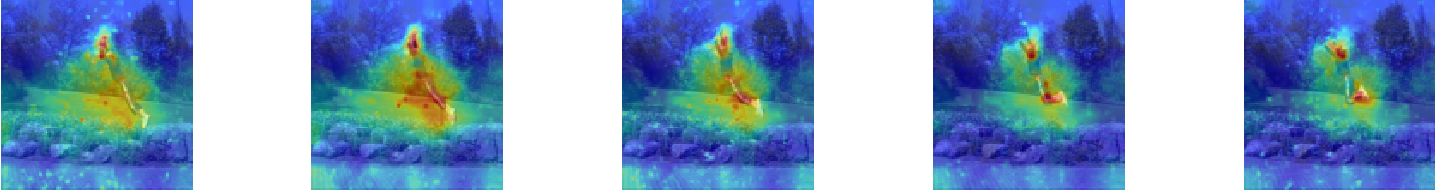}
\centering
\caption{\label{fig:temporal}Temporal information of 3D CNN using 3DCAM visual attribution. The saliency follows the object of interest over time.}
\end{figure*} 

Sometimes the critical information of the convolution layers' output activation is lost during propagation, starting from the input image or video to the last layer. The vanishing gradients occur when the information is far from the classification layer (fully connected layers). Hence, the last layer is considered suitable for representing the semantics of global-local interpretations. Thus, every layer's activation produces a higher attention mechanism with a higher probability of contributing pixels for the region of interest. Furthermore, the spatio-temporal poolings span vertically across convolution layers. It precludes consideration of using aggregated information between the top and bottom layers. Let us consider  $k$ given index representing layers with all 3DCAM layers aggregated. Then the nonlinear activation of pixel-wise ReLU is defined as

\begin{equation}
\textit{Sum3DCAM}=\textit{ReLU}\bigg(\sum_{k}{ 3DCAM_{k}}\bigg)
\end{equation}

Where Sum3DCAM has aggregated a set of 3DCAMs. Rectified Linear Unit (ReLU) activation is to suppress the negative weights into zero and highlight positive weight visualization points. In this study, as we are interested in every output convolution layer's salient part, the salient pixels are aggregated in a global-local manner. As shown in Figure \ref{fig:temporal} the top-down approach generates finer and higher sensitivity attribution on every input sequence.

\begin{figure*}
\centering
\includegraphics[width=1.0\textwidth]{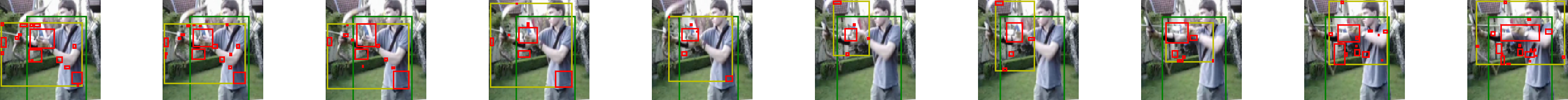}\\
\includegraphics[width=1.0\textwidth]{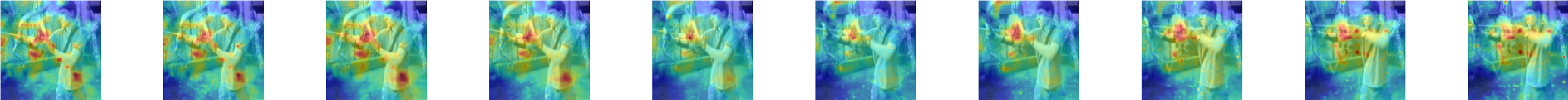}\\
\includegraphics[width=1.0\textwidth]{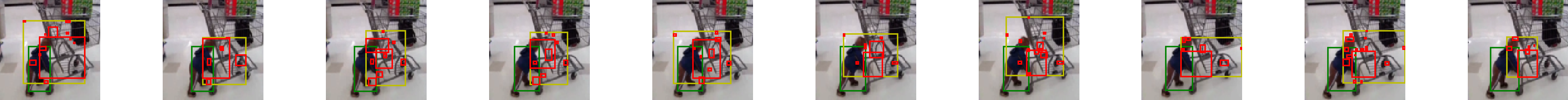}\\
\includegraphics[width=1.0\textwidth]{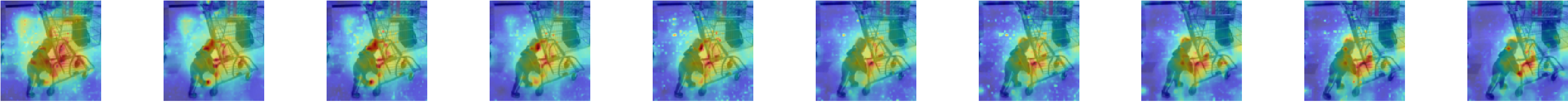}\\
\centering
\caption{\label{fig:bias} Visualization of 3DCAM which highlights actions in frames over time. The first and third row shows baby push walker. The second and fourth row shows saliency map visualization. It is shown that earlier frames tend to capture appearance as a whole and gradually focus on temporal saliency. It confirms the weight visualization done by C3D \cite{tran}.}
\end{figure*}

Furthermore, it is also noticed that the CNN filter captures the human motion together with the environment attached to it (Figure \ref{fig:bias}). Generally, the flow of information helps to enhance the salient weights. Similarly, in our proposed network with 3DCAM, the earlier frame captures spatial information and slowly becomes concentrated only on the salient (temporal) region. This visualization is concurrent with the investigation of C3D architecture\cite{tran}. 

\begin{figure}
\centering
\includegraphics[width=1.0\textwidth]{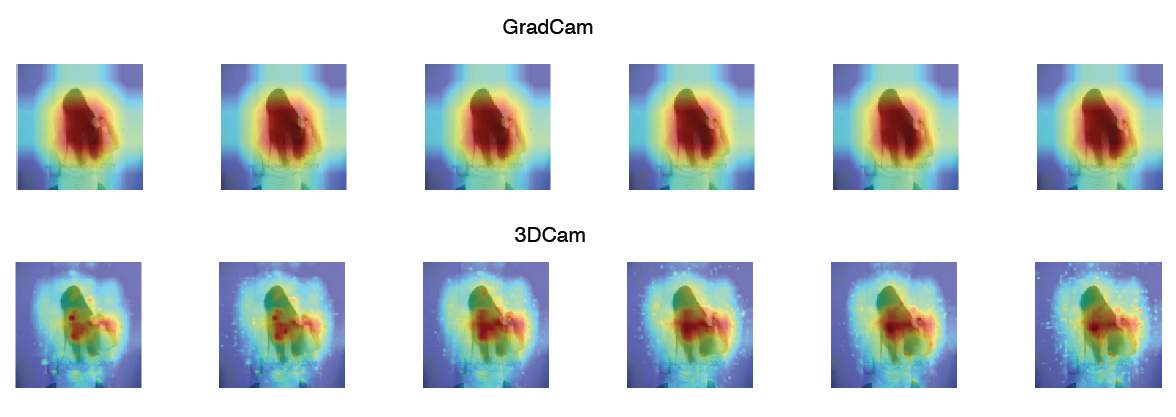}
\centering
\caption{\label{fig:vs}GradCAM vs 3DCAM}
\end{figure}

Our proposed approach shows high performance compared to GradCAM in capturing temporal information and fine-grained visualization, as shown in Figure \ref{fig:vs}. Moreover, it convincingly captures the most salient part of the image sequence by highlighting the hand action and movement rather than the whole body. In comparison, the GradCAM that considers the last or middle layer's attribution shows lower performance than ours.

\subsubsection{3DCAM Attributions to Sensitivity, Invariant, and Class-discriminative}
As shown in Figure \ref{fig:vs}, 3DCAM shows attributions that concentrate on specific regions compared to GradCAM. However, gradient-based attribution is said to be violating sensitivity, especially in the presence of the ReLU layer, because the gradients potentially become flat due to ReLU activation and thus make zero attribution \cite{shrikumar}. 
%If attribution is done on input rather than the last layer, it mainly occurs because there is long backpropagation through several convolution layers. It is even worse if only one convolution layer is employed and ReLU layer is applied at once.
However, the concept of implying the global-local aggregation with a discrete gradient can minimize the sensitivity problem. Therefore, 3DCAM is less sensitive due to underlying gradients. Let us consider, $g$ and $h$ are the first convolution layer and intermediate or final convolution layer, respectively, while $y$ is the final score. The aggregation of discrete gradients can be used in the form of $ \frac{\partial y}{\partial h}+\frac{\partial y}{\partial g}$. The more layers, the less sensitivity will likely occur. It is due to the complementary attributions of each layer.
Furthermore, the global-local aggregation used a total attribution score, similar to the backward chaining's attribution results. Hence, it is easy to implement invariants even though the discrete gradient is problematic if it neglects the intermediate layer and gathers the attribution score only from the input.

The proposed 3DCAM is class-discriminative, as shown in Figure \ref{fig:top_down}.  The first row shows the \emph{javelin throw} and the second row shows \emph{pole vault}. The proposed method attributes the javelin features but not the pole vault. It is because of the different features using the \emph{javelin throw} sequence as input and \emph{pole vault} as a baseline. Though the features are different, the proposed 3DCAM still achieves non-zero attribution in the \emph{javelin throw} sequence.    
%Implementation invariant means different networks with similar capability still show similar attribution. Backward chaining is related to implementation invariant since chaining nature means cover global and local attribution in a natural way (chain rule) and thus likely to be similar attribution will be given. 

%Additionally, it prevents attribution to be zero if in reality attribution should be present due to ReLU layer of which beckward chaining will have higher possibility of giving zero attribution.
\begin{figure}
\centering
\includegraphics[width=1.0\textwidth]{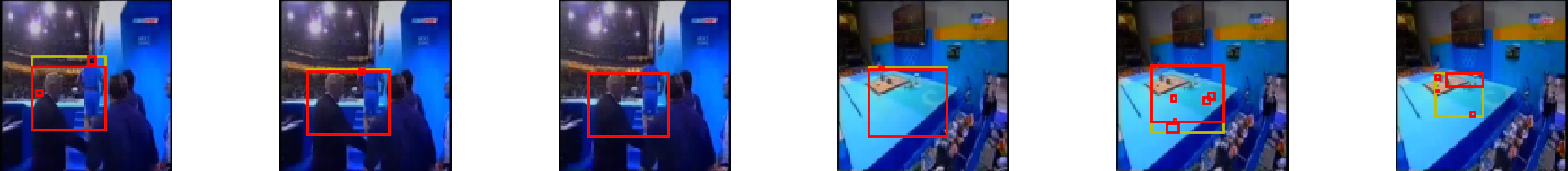}\\
\includegraphics[width=1.0\textwidth]{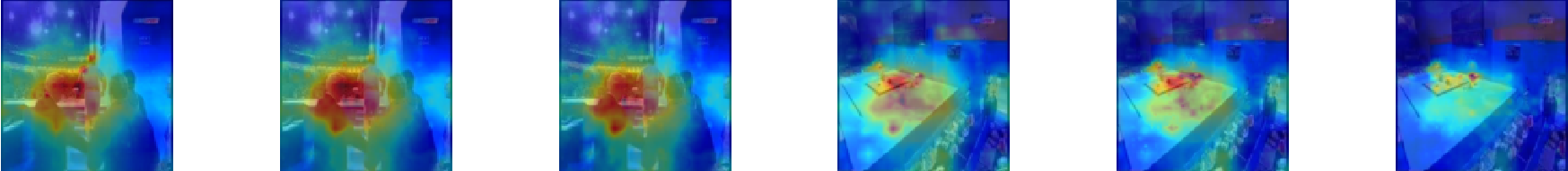}\\
\centering
\caption{\label{fig:limitations}The gradation of changes due to temporal upsampling. The scene transformation of untrimmed videos shows difficulties in finding temporal visualization.}
\end{figure}

\subsection{Limitations of 3DCAM}

In an untrimmed video, the temporal visualization is difficult due to its rough up-sampling of rigid and trilinear interpolation. If the latter scene is completely different from the former within the transition, then the visual attribution of 3DCAM finds it harder to differentiate between actions. The network weights are trained on the trimmed clips for specific tasks and categories. As shown in Figure \ref{fig:limitations}, the temporal visualization shows confusion in the middle of the transition between one action (weightlifting as a non-action part) and another one (billiard sport). It reveals that the weightlifting scene appears in two consecutive frames; even after the scene changes completely, it still occurs. Some transition samples (untrimmed with random sampling) should be trained or fine-tuned in the network that is trained on trimmed scenes to tackle this issue. In addition, it can be adapted to feature or meta-learning methods by detecting the scene transition over the temporal space. This strategy could be our future works.

\begin{figure}[h!]
\centering
\includegraphics[width=1.0\textwidth]{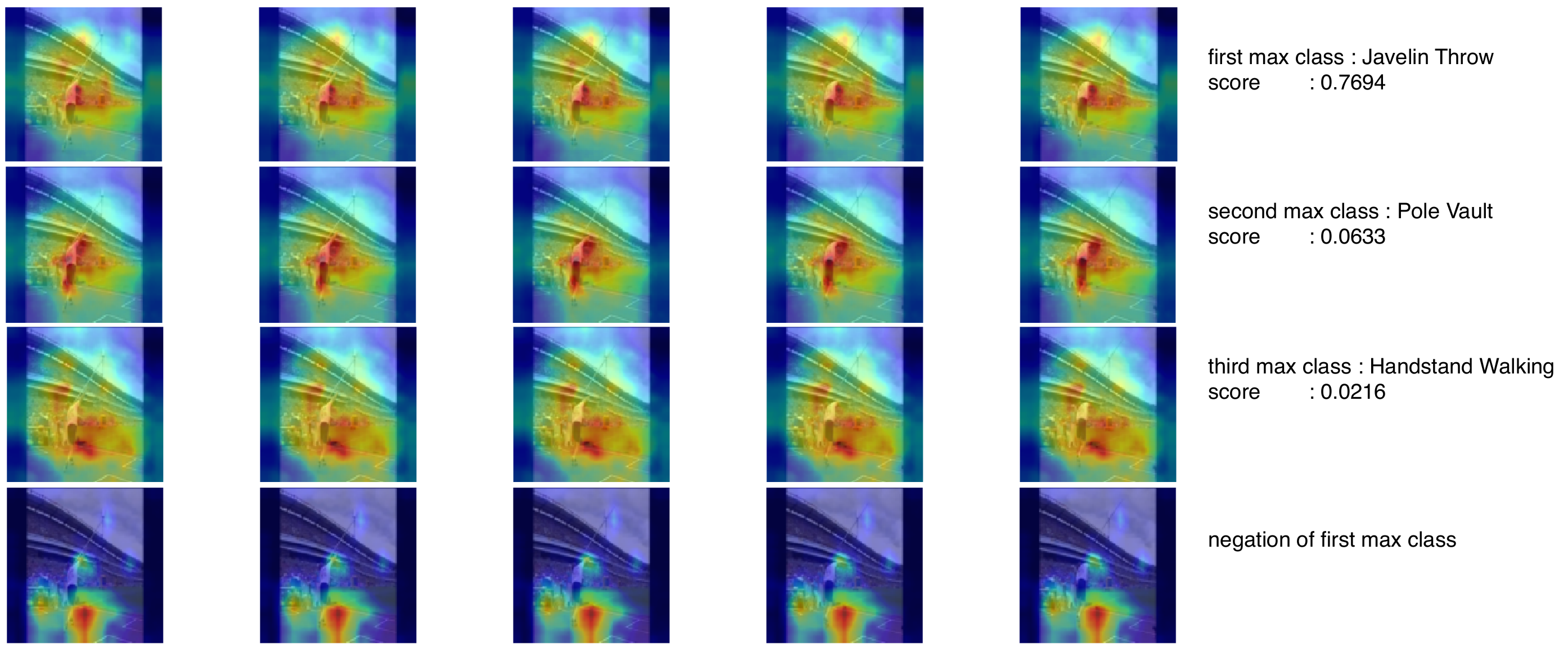}
\centering
\caption{\label{fig:top_down}Different visualization based on top three class scores and counterfactual explanation of maximum score.}
\end{figure}

%A number of previous works have asserted that deeper
%representations in a CNN capture higher-level visual constructs
%[5, 35]. 

%Furthermore, convolutional features naturally
%retain spatial information which is lost in fully-connected
%layers, so we can expect the last convolutional layers to
%have the best compromise between high-level semantics and
%detailed spatial information. 

%The neurons in these layers
%look for semantic class-specific information in the image
%(say object parts). 

%Grad-CAM uses the gradient information
%flowing into the last convolutional layer of the CNN
%to understand the importance of each neuron for a decision
%of interest. 

%Although our technique is very generic and can
%be used to visualize any activation in a deep network, in
%this work we focus on explaining decisions the network can
%possibly make.

\subsection{Counterfactual explanations using multi THUMOS dataset}

GradCAM is used to propose counterfactual explanations to test the region that supports the decision. This explanation negates the gradient of each layer, such that:

\begin{equation}
\hat{\beta_k}=
  \sum_{k}{ -\frac{\partial max(y)}{\partial \alpha_{ck}}}
\end{equation}

Where $\hat{\beta}$ is the negation of gradient field with respect to the maximum class score. Figure \ref{fig:top_down} demonstrates maximum score class (Javelin throw) and highlights javelin and the human body, while the second and third scores highlight only the human body area. It is found that several regions of the scene are the subset of the first score. Furthermore, the gradient's negation highlights the outer javelin and human body, which means 3D CNN can define the people's actions and objects related to that action.

\subsection{3DCAM on UCF24 dataset}

\begin{figure}[h!]
\centering
\includegraphics[width=1.0\textwidth]{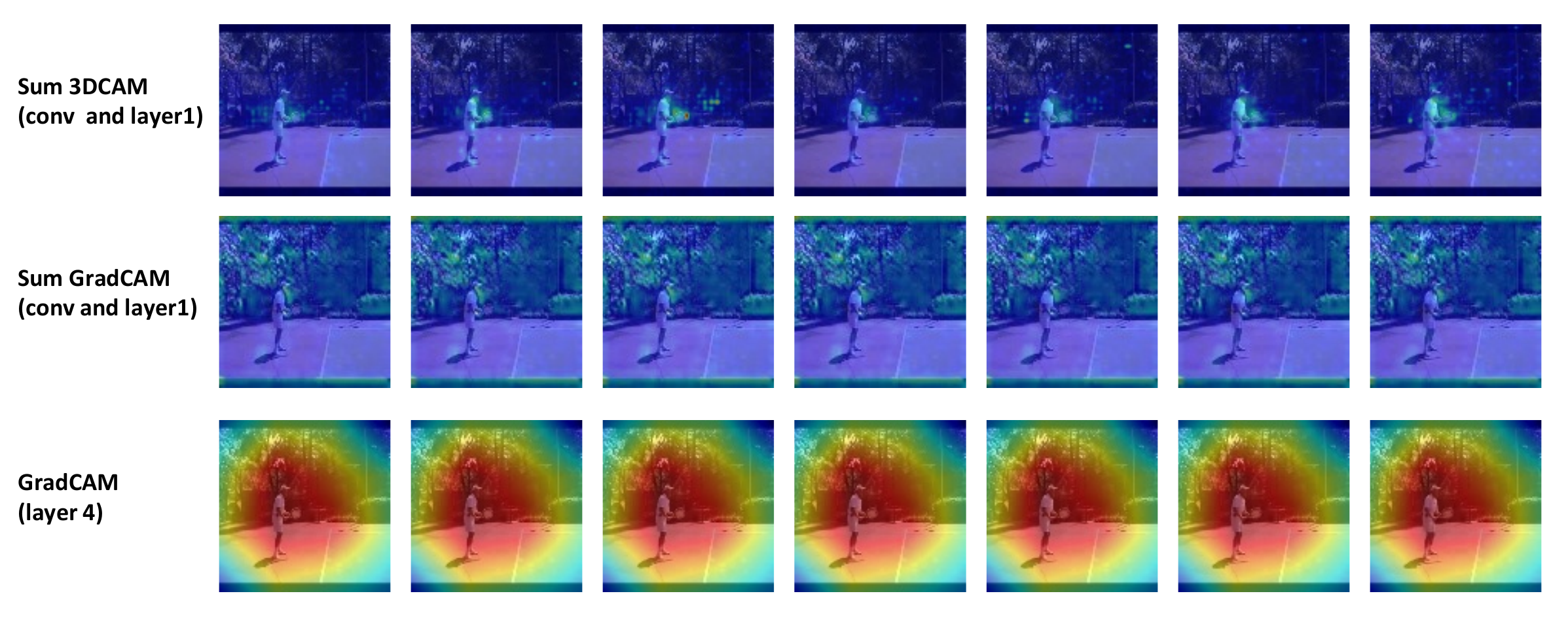}
\centering
\caption{\label{fig:3dCAMucf}Attention visualization of first to seventh frame of swing tennis from UCF24 dataset generated from 3DCAM and GradCAM.}
\end{figure}

As shown in Figure \ref{fig:3dCAMucf} the two upper layers (convolution and first residual layer) of 3D ResNext architecture generates finer and more localized attribution on every input sequence. Furthermore, it is also noticed that the CNN filter captures the human motion together with the tool attached to it (racket).
On UCF24 dataset, mostly, the size of actors is small compared to the multi THUMOS dataset. Therefore, upper layers that capture small features are suitable to be included in the sum of 3DCAM. On the other hand, as shown in Figures \ref{fig:3dCAMucf}, GradCAM of layer 4's attentions are too coarse due to global features, which is undesirable for weakly localization tasks. The choice of the upper layer is quantitatively confirmed in experiments and concurrent with IoU results for weakly localization tasks.

\subsection{Visual Attribution Network via 2DCAM on Images}

Our proposed global-local visual attribution is also tested on images via 2D CNN of ResNet50 architecture. 3DCAM can be reduced into 2DCAM by omitting temporal information and leaving spatial information. For upsampling, bilinear interpolation is used to adjust to 2D CNN.

\begin{figure}[H]
\centering
\includegraphics[width=1.0\textwidth]{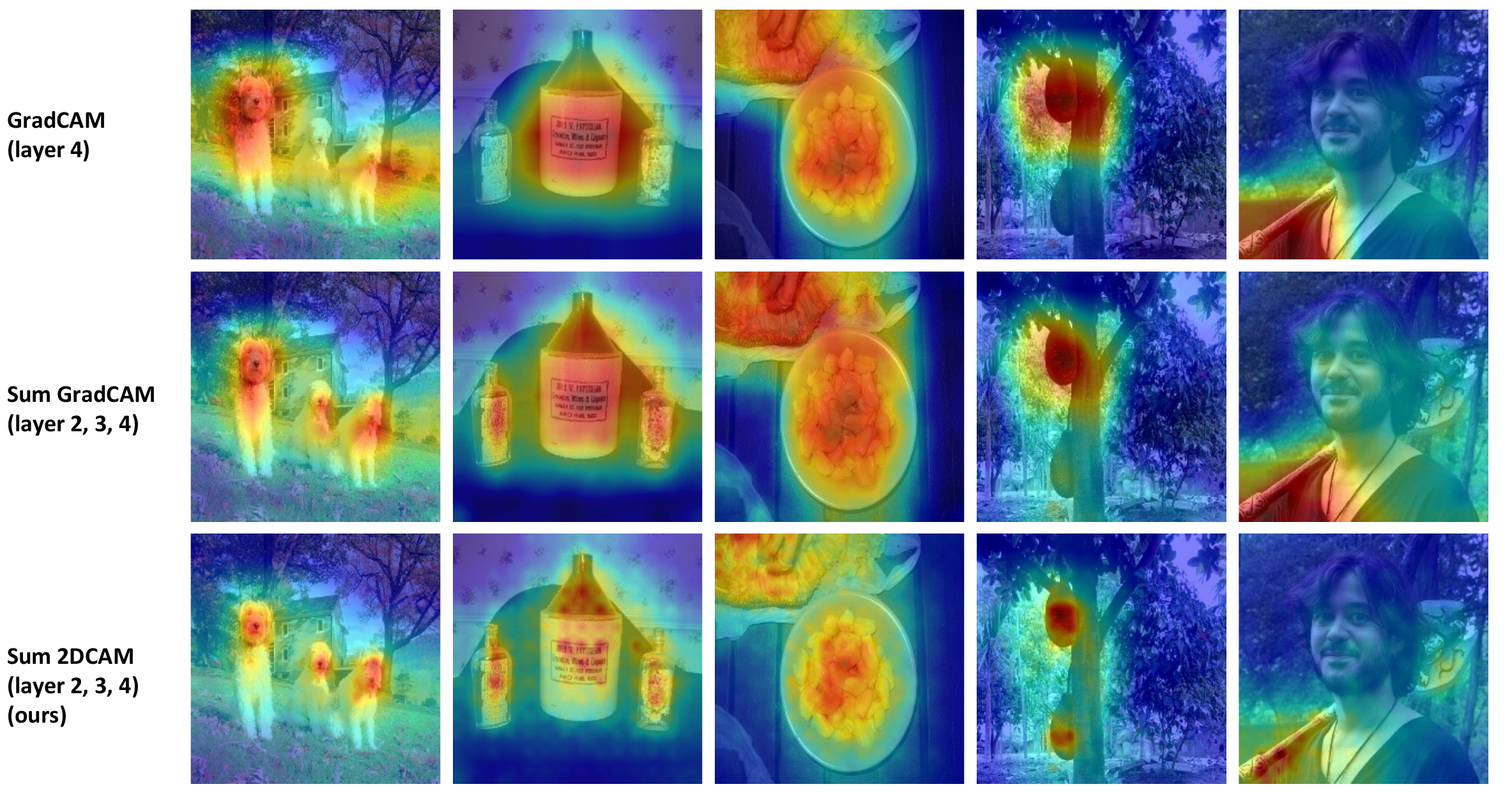}
\centering
\caption{\label{fig:2dcnn}Visual attention generated from GradCAM. Sum GradCAM, and Sum 2DCAM on ILSVRC dataset.}
\end{figure}

GradCAM works with deep CNN where the final $F(x)$ is a differentiable function of the activation maps $\alpha$, without any retraining or architectural modification. The GradCAM saliency maps are upsampled and fused in multilayers to obtain fine-grained pixel-scale representations.

As shown in Figure \ref{fig:2dcnn},  GradCAM fails to localize if the object of the same class appears more than one, as shown in the first row of 1-4 column images. This problem should be addressed properly in the real world, where most objects appear multiple times in an image. The global-local approach of Sum GradCAM generates finer localization, as shown in the second row of 1-4 column images. However, it still can not deliver distinctive attention between objects. GradCAM++ \cite{chattopadhay} uses weighted average of partial derivatives to address this localization issue. However, our direction is to incorporate global-local attention by summation of channels (Eqn. \ref{sumchannel}), proper ReLU, and normalization of each layer in the form of 2DCAM (Eqn. \ref{3DCAM}). This formulation, when accumulated in the form of Sum 2DCAM presents finer and distinctive localization of multiple objects, as shown in the last row of 1-4 column images. Moreover, it represents the object class in detail, as shown in the last row of the last column image of the axes class. This visualization is confirmed in experiments by comparison with GradCAM for weakly localization tasks via IoU thresholds to ILSVRC localization dataset.

\subsection{Action Recognition via Gated Attention}

Former investigations have proved that the depth provides essential information and can be applied to diverse recognition tasks \cite{szegedy}. In residual networks like ResNet, Kawaguchi et al. \cite{kawaguchi} theoretically showed that the depth and width improved local minima towards global minima. Furthermore, the local minima's quality is better than the global minima on the shallow network. It enhances the conclusion made by \cite{shamir} that it is little or no possibility of trained deep ResNet to achieve worse results than the shallower models. It regards local residual layers can confidently contribute to generalization and thus have richly important attention information.  

\begin{figure}
\centering
\includegraphics[width=1.0\textwidth]{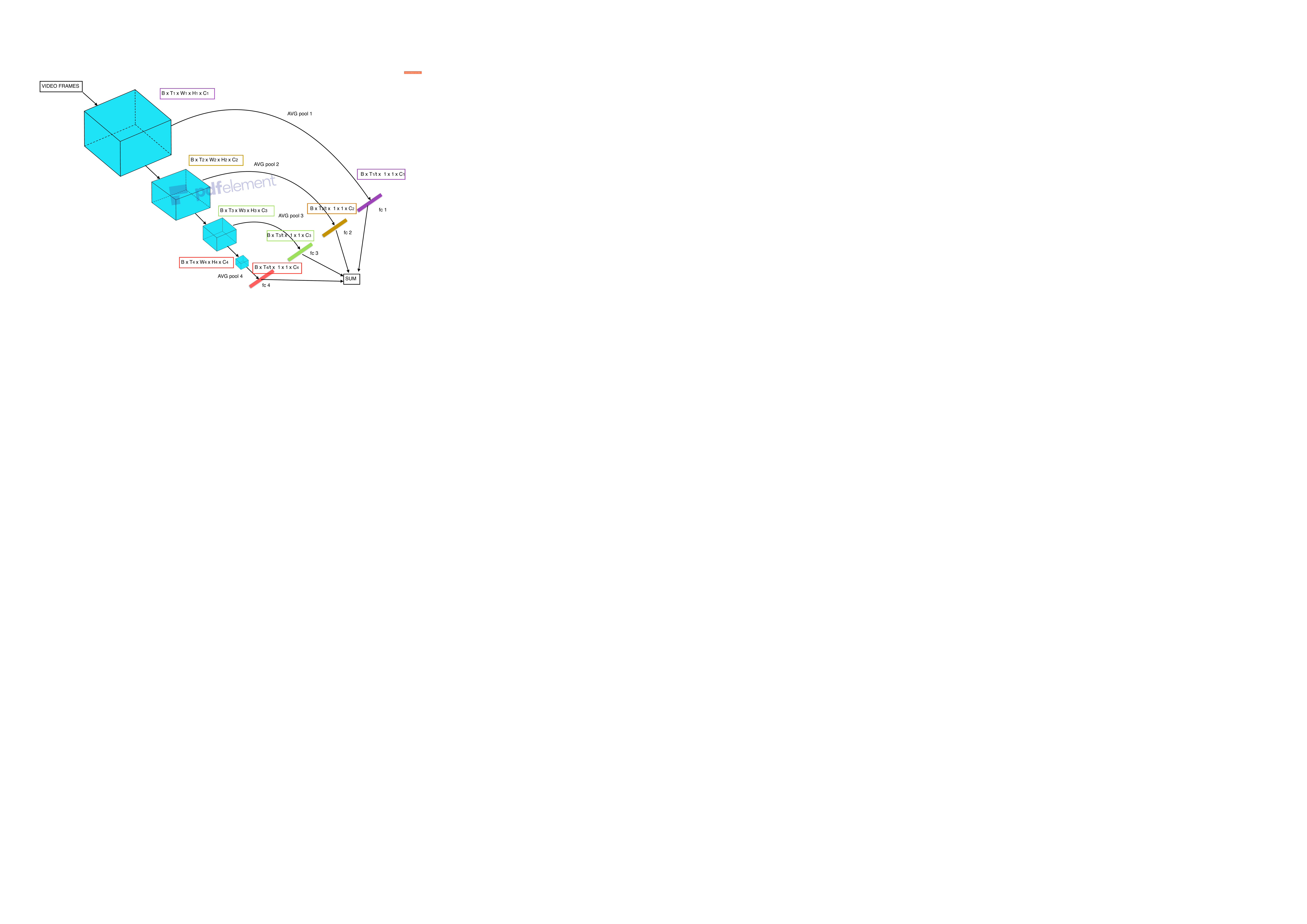}
\caption{\label{fig:3dgate}General scheme of spatio-temporal visual attribution of 3D CNN attention gate. }
\end{figure}

 Gated of each layer contributes to information, and thus it improves accuracy. Hence, our assumption is that if layer-wise attention contains valuable information, it can reveal important features through a fully connected layer (fc) that is lost during serial convolutions. We also use 3D ResNext to be gated in layer-wise for final action classification. Figure \ref{fig:3dgate} shows layer-wise gated architecture using layer-wise connection to fully connected (fc) layers for classification. For each targeted layer, the feature map is fed to the activation function of ReLU and pooled before the classification layer. The temporal channel is pooled from the temporal size of $T$ into the size of $T/t$ with factor of $t$ while spatial size of $W \times H \times C$ is pooled entirely. It brings a translation-invariant property to each layer while preventing temporal information. For instance, the activation map with size of $B \times T \times W \times H \times C$ is pooled into $B \times T/t \times 1 \times 1 \times 1$ before the fc layer. Gated mechanisms are implemented on selected layers.

\begin{comment}
\subsection{Do they have temporal information?}

In this subsection, we compare with optical flows (OF). OFs are subtracted by their mean to reduce CAMera motions. Intuitively, if 3DCAM has high overlapping (IoU) with OF, it has temporal motions that capture changes over time.

\begin{figure}
\includegraphics[width=1.0\textwidth]{optical_flows.png}\\
\includegraphics[width=1.0\textwidth]{viz_compare_flows.png}
\end{figure}
\end{comment}

\section{EXPERIMENTS}

\textbf{Dataset}. JHMDB provides per-frame bounding box annotations of joint annotated human performing actions. It includes 928 videos for 21 classes from the HMDB51 dataset \cite{kuehne}, and thus JHMDB is based on the HMDB51. HMDB51 is a trimmed video dataset containing 6,766 videos for 51 action classes. UCF101 \cite{soomro} provides trimmed action videos containing 13320 videos and 101 classes. It is usually used for action classification evaluation. Training/testing splits provided by this dataset guarantee fair performance evaluation by performing the same evaluation protocol. UCF24 is an action localization dataset derived from UCF101 dataset that contains 24 action classes. For 2D CNN, ILSVRC dataset, derived from ImageNet dataset, is used to evaluate weakly localization. ILSVRC dataset is a random subset of 50,000 images with labels used as validation data with the presence or absence of 1000 categories with box ground truth available.

\textbf{Models}. We use a 3D ResNext model by kenshohara \cite{hara} which achieved  63.8\% on HMDB51. Initially, 3D ResNext is pre-trained on Kinetics dataset before fine-tuned on HMDB51. The trained weights are then used for visual attribution. Note that we sample a non-overlapping clip that contains 16 frames in sequence for all videos. We also use 3D ResNext in layer-wise with fc layers for action classification. As shown in table \ref{3dresnext}, within 3D ResNext, the output of conv3d 1a, mixed 2, mixed 3, mixed 4, and mixed 5 are connected to their respective fc layer in the form of gated connection. The $t$ factor is set to two, giving equal temporal division as the network goes deeper. On UCF24 evaluation, we used ResNext model with transfer learning from Kinetics-400 and fine-tuned on UCF-101 dataset. The data is sampled at a sampling rate of 1 in the amount of 16 frames at every sampling. The prediction boxes are generated from videos where the ground truth boxes are present. Besides 3D ResNext, we also prepare I3D \cite{carreira} which is pre-trained on Kinetics, delivering state-of-the-art accuracy in action classification. As shown in table \ref{i3d}, I3D has 11 layers which are mostly mixed convolution layers. Considering simplicity, the gating connection is set to the output of mixed 3d, mixed 4f, and mixed 5c. For 2DCAM, we used ResNet50 with transfer learning from ImageNet to predict the class of ILSVRC dataset \cite{olga}. For 2D SNN, we used ResNet-50 model with transfer learning from ImageNet for the classification task.

\textbf{Hit point}. Noticing that visualization is not intentionally optimized for localization, thus hit point metric is a suitable measurement test. The test is performed by C-mwp \cite{zhang} and GradCAM \cite{selv}. The hit point uses the formulation of  $Acc = \#Hits/(\#Hits + \#Misses)$, where if the saliency intersects the bounding box, it is termed to be $Hit$. $Acc2 = c\times\#Hits/(\#Hits + \#Misses)$ where $c$ is 1 if the prediction class is true and 0 otherwise for the purpose of visual attribution. The hit point is the same as $Acc2$ if the IoU (intersection over union) threshold is 0.0.

\textbf{MaxBoxAcc}. We follow the evaluation of \cite{chloe} by defining the box accuracy at score map threshold $\tau$ and IoU threshold $\theta$. MaxBoxAcc is defined as the tightest boxes surrounding mask. IoU of boxes A and B is the maximum value of a set of IoU between boxes A and B. The performances are then averaged across threshold $\theta \in $ \{0.3,0.5, 0.7\} for ILSVRC localization dataset and $\theta \in \{0.05,0.1, 0.2, 0.3,0.4,0.5\}$ for UCF24 action localization dataset. To accommodate multi-contour thresholding, we used contour thresholding of $\tau$ with values ranging from 0 to 1 with a step of 0.01 to generate multiple boxes possible as attention. For UCF24 dataset, we choose contour generated from $\tau$, which produces maximum IoU, while on ILSVRC dataset, we use all contours generated from a set of $\tau$.

\textbf{Hyperparameters}. For action classification, we use an initial learning rate of 0.001, decreasing by multiplying by 0.1 for every 50 epochs. Stochastic Gradient Descent (SGD) is used as an optimization algorithm. Regardless of the network architecture, the same hyperparameters are set in all experiments.

\section{Results and Discussions}
\subsection{Performance Detection Based on Visual Attribution}

In this experiment, we back-propagated the decision made by the 3D model (3D ResNext). In order to evaluate the visual attribution, we compare the proposed approach with the baseline visual attribution of GradCAM. We evaluate untrimmed weakly localization in both spatial and time-space due to the difficulties in finding the untrimmed dataset that includes bounding box localization and is not intentionally optimized for localization. Finally, we validate our grounding results with ground-truth bounding boxes that localize actions per frame.

\begin{table}[H]
\centering
\scalebox{0.9}{
\begin{tabular}{| c | c | c|}
\hline
\textbf{Convolutions} &  \textbf{Output size ($T\times H \times W$)} & \textbf{Layers} 
\\
\hline
Conv3d 1a $7\times7$  & $8\times112 \times 112$ & conv \\
MaxPool3d 2a $3\times3$  && - \\
Mixed 2  & $8\times56\times56$ & layer 1 \\
Mixed 3  & $4\times28\times28$ & layer 2 \\
Mixed 4 & $2\times14\times14$ & layer 3
\\
Mixed 5 & $1\times7\times7$ & layer 4
\\

\hline
\end{tabular}
}
\caption{\label{3dresnext}3D ResNext-101 layers}
\end{table}

\begin{table}[H]
\centering
\scalebox{0.9}{
\begin{tabular}{| c | c | c|}
\hline
\textbf{Convolutions} &  \textbf{Output size ($T\times H \times W$)} & \textbf{Layers} \\
\hline
Conv3d 1a $7\times7$  && conv 1 \\
MaxPool3d 2a $3\times3$  && - \\
Conv3d 2b $1\times1$  &&  \\
Conv3d 2c $3\times3$  &&  \\
MaxPool3d 3a $3\times3$ &&
\\
Mixed 3b && layer 1
\\
Mixed 3c & $32\times28\times28$ &
\\
MaxPool3d 4a $3\times3$ && \\
Mixed 4b && layer 2\\
Mixed 4c && \\
Mixed 4d && \\
Mixed 4e && \\
Mixed 4f & $16\times14\times14$ & \\
MaxPool3d 5a $2\times2$ && \\
Mixed 5b && layer 3\\
Mixed 5c & $8\times7\times7$ & \\
Logits &&\\
\hline
\end{tabular}
}
\caption{\label{i3d}I3D layers}
\end{table}

\begin{table}[H]
\centering
\scalebox{0.9}{
\begin{tabular}{| c | c |}
\hline
\textbf{Method} &  \textbf{Accuracy} 
\\
\hline
GradCAM (last layer) \cite{selv}  & 0.988 \\
First conv layer  & 0.821 \\
Second conv layer  & 0.956 \\
Third conv layer  & 0.951 \\
Fourth conv layer & 0.825
\\
Fifth conv layer & 0.976
\\
%average pool layer & \bf0.992
%\\
All layers & \bf0.991\\
\hline
\end{tabular}
}
\caption{\label{acclayers}Estimation of accuracy based on pointing game (hit point) on JHMDB dataset}
\end{table}

\begin{table}[H]
\centering
\scalebox{0.9}{
\begin{tabular}{|*{3}{c|}}
\hline

\multicolumn{1}{|c|}
{\textbf{Method}}&\multicolumn{2}{c|}{IoU threshold}\\
\hline
&0.5&0.6\\
\hline
GradCAM  \cite{selv}   &0.08&0.013 \\
Avg pool 3DCAM   & 0.089&0.008 \\
Sum 3DCAM  & \bf0.127 & \bf0.05 \\
\hline
\end{tabular}
}
\caption{\label{acciou}Estimation of accuracy based on the overlapping of the bounding boxes on JHMDB dataset with the given IoU threshold}
\end{table}

\begin{table}[H]
\centering
\scalebox{0.9}{
\begin{tabular}{|*{3}{c|}}
\hline
\multicolumn {1}{|c|}{Method}&\multicolumn{2}{c|}{IoU threshold}\\
\hline
&0.5&0.6\\
\hline
GradCAM  \cite{selv}   & 0.052&0.003 \\
Avg pool 3DCAM   & 0.052& \bf0.052 \\
Sum 3DCAM  & \bf0.076 &0.03 \\

\hline
\end{tabular}
}
\caption{\label{acc2iou}Estimation of accuracy (Acc2) based on the classification and localization on JHMDB dataset with the given IoU threshold}
\end{table}

\begin{comment}
\begin{table}[H]
\centering
\scalebox{0.9}{

\begin{tabular}{| c | c | c | c | c | c | c | c|}
\hline
\textbf{Method} &  \textbf{Hit}&
\textbf{0.1}&
\textbf{0.2}&
\textbf{0.3}&
\textbf{0.4}&
\textbf{0.5}&
\textbf{0.6}
\\
\hline
GradCAM \cite{selv}   & 0.99 &0.918&0.764&0.535&0.303&0.115&0.019 \\
sum 3DCAM  & \bf0.991 & \bf0.92 & 0.72& 0.474& 0.269 &\bf 0.131 & \bf 0.052 \\

\hline
\end{tabular}
}
\caption{Precision of Localization on JHMDB dataset}
\end{table}
\end{comment}

Table \ref{acclayers} shows each layer's ablation study. It shows that the average pool gives results similar to all summed 3DCAM. In the JHMDB dataset, humans are mostly located across various scales in the center of the frames. As shown in Table \ref{acciou}, there is no significant difference when using a threshold of 0.0 (the same as the hit point). However, when it comes to the more detailed IoU threshold of 0.5 and 0.6, 3DCAM shows an intersection improvement of 58.75 \% and 284.61  \%, respectively, over GradCAM. Table \ref{acc2iou} shows similar results of no significant difference in acc2 when using a threshold of 0.0. However, when it comes to a more detailed threshold of 0.5 and 0.6, 3DCAM outperforms intersection improvement of 46.15 \% and 90.0  \%, respectively, over GradCAM.

Furthermore, the proposed 3DCAM shows high saliency on some body parts rather than its full body. Thus weak localization tends to localize a subset of the full human body. Besides, we found that it points to the objects attached to human action. Hence parts of the human body and objects attached to the actions can be included as factors in the decision. For example, as shown in Figure 3, the "baby pushing the table" action attributes the head, hands of the baby and the table.

\subsection{Performance of Weakly Localization based on Visual Attribution on UCF24}

\begin{table}[H]
\centering
\scalebox{0.8}{
\begin{tabular}{|*{8}{c|}}
\hline
\multicolumn {1}{|c|}{Method}&\multicolumn{7}{c|}{IoU threshold}\\
\hline
&0.05&0.1&0.2&0.3&0.4&0.5&average\\
\hline

3DCAM conv   & 0.225
&0.211
&0.173
&0.133
& 0.118
& \bf0.090
&0.159\\
3DCAM layer 1    & 0.827
&0.651&0.38
&0.196
&0.099
&0.044
&0.366\\
3DCAM layer 2     & 0.826
&0.651
&0.375
&0.19
& 0.09
& 0.044&0.364
\\
3DCAM layer 3    & 0.828
&0.651&0.373&0.19& 0.090& 0.038&0.365\\
3DCAM layer 4    & 0.826&0.651&0.373&0.191& 0.090& 0.038&0.361\\
Sum 3DCAM  (layer conv and 1) (ours) & \bf0.835 &\bf0.676& \bf0.426&\bf0.238&\bf0.132&0.063& \bf0.395\\
Sum 3DCAM (layer conv, 1,2,3) (ours)  & 0.826
&0.651
&0.378
&0.215
& 0.099
& 0.051
&0.37\\
Sum GradCAM (layer conv and 1)  \cite{selv}  & 0.826&0.651&0.373&0.191& 0.090& 0.038&0.361\\
\hline
\end{tabular}
}
\caption{\label{maxvucf24}Estimation of accuracy (MaxBoxAcc) based on the  localization on UCF24 with the given IoU thresholds}
\end{table}

Table \ref{maxvucf24} shows each layer's ablation study of 3DCAM, GradCAM, and its aggregation. Overall, It shows that Sum 3DCAM produces localization with IoU thresholds (0.05,0.1.0.2.0.3,0.4, and 0.5) which is better than 3DCAM of a layer except for the IoU threshold of 0.5 where 3DCAM is generated from the top of the convolution layer. Note that, in the UCF24 dataset, the actors appear relatively small compared to the size of the video, as can be seen from the visualization of the swing tennis class in Figure \ref{fig:3dCAMucf}. Because the size of the ground truth localization box in the UCF24 dataset is, on average small and detailed, the top layers are suitable for capturing attention through 3DCAM. Based on the average of IoU thresholds, 3DCAM with aggregation of the top layer (convolution) and layer1 obtains an average MaxAccBox of $39.5 \%$. This is better than Sum GradCAM with the same layer, namely the convolution layer and layer 1, with a difference of $2.5\%$. There is no significant difference in the average of MaxAccBox accuracy among 3DCAM using a layer except for the topmost layer, namely the convolution layer, which on average produces only $15.9 \%$. However, 3DCAM at the topmost convolution layer produces the best  MaxAccBox at an IoU threshold of 0.5. This shows that global-local aggregation in weakly localization actions is important. However, we need to choose which layers are suitable based on the scale of actors related to the dimensions of the whole video.

Furthermore, when compared to GradCAM, 3DCAM shows a cleaner and more detailed attention localization, as shown in Figure \ref{fig:3dCAMucf}. Based on the visualization, it appears that 3DCAM captures the actors and their equipment over time in the video. However, if we look at GradCAM with the same layer composition, there are false positives attentions which can lead to worse localization than GradCAM.

\subsection{Performance of Weakly Localization based on Visual Attribution on 2DCNN}

\begin{table}[h!]
\centering
\scalebox{0.9}{
\begin{tabular}{|*{5}{c|}}
\hline
\multicolumn {1}{|c|}{Method}&\multicolumn{4}{c|}{IoU threshold}\\
\hline
&0.3&0.5&0.7&average\\
\hline
GradCAM (layer 4)  \cite{selv}   & 0.803& 0.589& 0.317&0.57\\
Sum GradCAM (layer 2,3,4))  \cite{selv}   & 0.79& 0.622& 0.335&0.582\\
Sum 3DCAM (layer 2,3,4)  & \bf0.819& \bf0.622& \bf0.354& \bf0.598\\
\hline
\end{tabular}
}
\caption{\label{2dcnnacc}MaxBoxAcc accuracy of weakly localization on ILSVRC localization dataset given IoU threshold}
\end{table}

Table \ref{2dcnnacc} shows MaxBoxAcc accuracy using several IoUs (0.3, 0.5, and 0.7). Based on the table, across IoU thresholds,  Sum 3DCAM achieves the highest matching frequency with an average MaxBoxAcc of $59.8 \%$. The highest accuracy is performed by Sum GradCAM  ( layer1, 2, and 3 ) at an IoU threshold of 0.5 of $62.2 \%$, which is the same as Sum GradCAM. However, Sum 3DCAM outperformed Sum GradCAM at IoU thresholds of 0.3 and 0.7 with a difference of $2.9\%$ and $1.9 \%$, respectively. When compared to GradCAM of layer 4 only, it shows that the aggregation of 3DCAM of layers 2,3, and 4 in the form of Sum GradCAM outperformed with differences of $1.6 \%$, $3.3 \%$, $3.7\%$ at IoU threshold of 0.3, 0.4, 0.4, respectively. This shows that information from several layers, from global to local, is needed to capture more detailed attention. This can be visually validated in Figure \ref{fig:2dcnn} which shows that aggregation of 3DCAM of layers 2,3, and 4 can increase attentional detail for several object classes. Moreover, it generates better multiple object localization. Furthermore, Sum 3DCAM shows more detailed and precise attention to objects than Sum GradCAM. This is correlated with the quantitative evaluation results as shown in the table \ref{2dcnnacc}.

\begin{table}[h!]
\centering
\scalebox{0.9}{
\begin{tabular}{|*{3}{c|}}
\hline
\multicolumn {1}{|c|}{Method}&\multicolumn{2}{c|}{IoU threshold}\\
\hline
&0.5&0.6\\
\hline
GradCAM  \cite{selv}   & 0.052&0.003 \\
Sum 3DCAM  & \bf0.076 &\bf0.03 \\

\hline
\end{tabular}
}
\caption{\label{acc2iou}Estimation of accuracy (Acc2) based on the classification and localization on JHMDB dataset with the given IoU threshold}
\end{table}

\subsection{Performance of Layer-wise Gated 3D CNN on Action Recognition}

\begin{table}[H]
\centering
\scalebox{0.9}{
\begin{tabular}{| c | c | c | c | c | c | c | c|}
\hline
\textbf{Method} & \textbf {modality}& \textbf{frame-wise}&
\textbf{total video-wise}
\\
\hline
Resnext-101   & rgb&0.875&0.891\\
Resnext-101 (layer 4+3+2+1+conv) &rgb&\bf0.895&\bf0.919 \\

\hline
\end{tabular}
}
\caption{\label{3dresnextcls}Estimation of incremental accuracy using 3D ResNext-101 network}
\end{table}

\begin{table}[H]
\centering
\scalebox{0.9}{
\begin{tabular}{| c | c | c | c | c | c | c | c|}
\hline
\textbf{Method} & \textbf{modality}&\textbf{accuracy}
\\
\hline
I3D & rgb &0.962 \\
I3D ( layer 3)&rgb   & 0.968\\
I3D ( layer 2)&rgb   & 0.967\\
I3D ( layer 1+2+3)&rgb   & 0.961\\
I3D ( layer 2+3)& rgb & \bf0.971 \\
I3D ( layer 3)& rgb + flow & 0.978\\
\hline
\end{tabular}
}
\caption{\label{i3dcls}Estimation of incremental accuracy using I3D network}
\end{table}

In action recognition, we compare gated architecture with its baseline. Table \ref{3dresnextcls} presents our proposed gated version of 3d ResNext produces higher accuracy of 2.8\% than the baseline 3D ResNext in total video-wise recognition. In table \ref{i3dcls} we can find that the gated version of the I3D network (RGB mode) is better (0.9 \%) than the baseline I3D.  The best accuracy's method (I3D with RGB and Flow modalities) includes more number parameters due to its multi-modality structure. In our gated network, even with an uni-modality network (RGB mode) reveals a global-local gating mechanism has fewer parameters and thus more efficient than the multi-modality network (only has difference of -0.7\%). Therefore, when applied to action classification in the form of attention gated network, it improves the network's accuracy and marks the importance of global-local attention.

\section{CONCLUSIONS}

This study highlights the importance of global-local visual explanation, weakly-supervised localization, and action recognition in 3D CNN. We proposed 3DCAM that efficiently class-discriminated actions's attribution and produced high resolution spatio-temporal visual explanations. The proposed approach efficiently recovered the spatio-temporal information, which is lost during convolutional and pooling mechanism in the baseline 3D-based models. Experimental results based on weakly-supervised localization analysis showed that 3DCAM outperforms the baseline GradCAM with a more detailed visual explanation in terms of hit point and IoU measures. Furthermore, the global-local mechanism in the gating network for action classification contributed to improvement of accuracy than the observed baseline. Based on the results of weakly-supervised action localization, temporal action detection or visual explanation, and action recognition, we suggest that the global-local attention could be an useful tool for many computer vision tasks in more diverse domains that use 3D CNN as a backbone.

\section*{ACKNOWLEDGMENT}

The authors would like to thank KAKENHI project no. 16K00239 for funding the research.

\end{document}